\pdfoutput=1

\documentclass[11pt]{article}

\usepackage{acl}

\usepackage{times}
\usepackage{latexsym}

\usepackage[T1]{fontenc}
\usepackage[utf8]{inputenc}
\usepackage{microtype}
\usepackage{xspace}
\usepackage{inconsolata}

\usepackage{xcolor}
\definecolor{codegreen}{rgb}{0,0.6,0}
\definecolor{codegray}{rgb}{0.5,0.5,0.5}
\definecolor{codepurple}{rgb}{0.58,0,0.82}
\definecolor{backcolour}{rgb}{0.95,0.95,0.92}
\usepackage{listings}
\lstdefinestyle{mystyle}{
    backgroundcolor=\color{backcolour},   
    commentstyle=\color{codegreen},
    keywordstyle=\color{magenta},
    numberstyle=\tiny\color{codegray},
    stringstyle=\color{codepurple},
    basicstyle=\ttfamily\footnotesize,
    breakatwhitespace=false,         
    breaklines=true,                 
    captionpos=b,                    
    keepspaces=true,                 
    numbers=left,                    
    numbersep=5pt,                  
    showspaces=false,                
    showstringspaces=false,
    showtabs=false,                  
    tabsize=2
}
\usepackage{graphicx}
\usepackage{multirow}
\usepackage{booktabs,xltabular}
\usepackage[ruled,vlined,algo2e]{algorithm2e}
\usepackage{algorithm}
\usepackage{algorithmic}
\usepackage{enumitem}
\usepackage{caption}
\let\Algorithm\algorithm
\renewcommand\algorithm[1][]{\Algorithm[#1]\setstretch{1}}

\title{Decoding on Graphs: Faithful and Sound Reasoning on Knowledge Graphs through Generation of Well-Formed Chains}

\author{Kun Li$^{\heartsuit}$\thanks{$\;\;$Equal contribution.}$\;\,$, Tianhua Zhang$^{\heartsuit*}$, Xixin Wu$^{\heartsuit}$, \\ \bf Hongyin Luo$^{\diamondsuit}$, James Glass$^{\diamondsuit}$, Helen Meng$^{\heartsuit‡}$\thanks{$\;\;$Corresponding author.} \\
$^\heartsuit$The Chinese University of Hong Kong, Hong Kong SAR, China \\
$^\diamondsuit$Massachusetts Institute of Technology, Cambridge MA, USA \\
\texttt{kunli@se.cuhk.edu.hk, thzhang@link.cuhk.edu.hk}
}

\begin{document}
\maketitle
\begin{abstract}
Knowledge Graphs (KGs) can serve as reliable knowledge sources for question answering (QA) due to their structured representation of knowledge. Existing research on the utilization of KG for large language models (LLMs) prevalently relies on subgraph retriever or iterative prompting, overlooking the potential synergy of LLMs' step-wise reasoning capabilities and KGs' structural nature. In this paper, we present \text{DoG} (\textbf{D}ecoding \textbf{o}n \textbf{G}raphs), a novel framework that facilitates a deep synergy between LLMs and KGs. We first define a concept, \textit{well-formed chain}, which consists of a sequence of interrelated fact triplets on the KGs, starting from question entities and leading to answers. We argue that this concept can serve as a principle for making faithful and sound reasoning for KGQA. To enable LLMs to generate well-formed chains, we propose \textit{graph-aware constrained decoding}, in which a constraint derived from the topology of the KG regulates the decoding process of the LLMs. This constrained decoding method ensures the generation of well-formed chains while making full use of the step-wise reasoning capabilities of LLMs. Based on the above, \textsc{DoG}, a training-free approach, is able to provide faithful and sound reasoning trajectories grounded on the KGs. Experiments across various KGQA tasks with different background KGs demonstrate that \textsc{DoG} achieves superior and robust performance. \textsc{DoG} also shows general applicability with various open-source LLMs.

\end{abstract}

\section{Introduction}
Large language models (LLMs) have shown impressive performance across various natural language processing tasks \cite{NEURIPS2020_1457c0d6, ouyang2022traininglanguagemodelsfollow, Achiam2023GPT4TR, dubey2024llama3herdmodels}. Despite this, LLMs still suffer from the lack of knowledge and are prone to hallucinations
for some knowledge-intensive scenarios \cite{yin-etal-2023-large, hong-etal-2023-faithful}. One direction to addressing this limitation is to supplement LLMs with external knowledge \cite{10.5555/3495724.3496517, li-etal-2022-grounded, asai2024selfrag}. Among the external knowledge sources, knowledge graphs (KGs) lend themselves to knowledge-intensive tasks like question-answering, due to their structured and explicit representations of knowledge \cite{sun2024thinkongraph, jiang-etal-2023-structgpt, baek-etal-2023-knowledge}.


Many researchers have studied the augmentation of KGs as knowledge sources for LLMs when handling question answering (QA) tasks \cite{sun2024thinkongraph, luo2024reasoning, markowitz-etal-2024-tree, he2024gretrieverretrievalaugmentedgenerationtextual, wang2023knowledgedrivencotexploringfaithful}. These KGQA approaches can be mainly categorized into the following two threads. (1) \citet{luo2024reasoning}, \citet{he2024gretrieverretrievalaugmentedgenerationtextual} and \citet{mavromatis2024gnnraggraphneuralretrieval} rely on specialized subgraph retrievers to retrieve question-focused subgraphs from a complete KG, which are then fed into LLMs for predicting answers. Such approaches shift the burden of finding answers on KGs to the subgraph retrievers, while the LLMs merely select the most possible answer from the highly concentrated subgraphs and are not deeply involved in the graph reasoning process. This only makes limited use of LLMs' reasoning capabilities. On the other hand, training a specialized subgraph retriever demands substantial labeled data, and the trained retriever may struggle with out-of-domain scenarios (a retriever trained on one KG can not adapt to other KGs with disparate taxonomies, \textit{e.g.}, Freebase \cite{10.1145/1376616.1376746} \textit{vs.} Wikidata \cite{Vrandei2014Wikidata}, §\ref{section_main_results}). (2) To address these shortcomings, the second thread of works directly engages LLMs in the reasoning process on KGs \cite{sun2024thinkongraph, markowitz-etal-2024-tree, sun-etal-2024-oda}, through iteratively prompting LLMs to generate a series of operations on KGs (\textit{e.g.}, neighbor entities exploration, path decision), finally reaching to an answer. Such approaches, however, place significant demands on the LLMs' abilities, since the instructions intended for some complex operations are often difficult for small-sized LLMs (with < 10 billion parameters) to understand and execute, as evidenced by the fact that the backbone LLMs used in these works all have tremendous sizes, \textit{e.g.}, GPT-4, Llama 2-70b. 

Recently, research on chain-of-thought (CoT, \citeauthor{wei2022chain}, \citeyear{wei2022chain}, \citeauthor{feng2023towards}, \citeyear{feng2023towards}) has established that LLMs can solve some complex questions through the generation of step-by-step reasoning trajectories. Meanwhile, the structured nature of KGs, in which related facts are connected (directly or indirectly) through paths represented by a sequence of relations and entities between them, facilitates sound reasoning trajectories on graphs. In that sense, a tight coupling between LLMs' step-wise reasoning capability and KGs can be a promising solution to KGQA. In this work, building upon these considerations, we explore a perspective different from the aforementioned two threads of works: \textit{let LLMs reason directly on KGs through generation of reasoning steps}. Specifically, given a KG as input, the LLM is required to reason on KGs by generating sequential and interrelated reasoning steps, with each step being anchored by a fact triplet on the KGs. This manner can make full use of LLMs' textual understanding and generation capabilities for discovering suitable paths from question to answer, leading to general applicability to different KGs. 
Moreover, predicting the next token is easier than following complex instructions for small LLMs.

Motivated by this, we first define a concept, \textbf{\textit{well-formed chain}}, to serve as a principle for making faithful and sound reasoning on KGs. A well-formed chain, starting from query entities and leading to answers, is composed of a sequence of interconnected triplets as the reasoning steps, and each step of triplet can only grow from all previously visited triplets. Thanks to its properties, for a question, a well-formed chain offers a reasoning trajectory that is sound and faithful to the KG; Moreover, it will also naturally narrow down the search scopes for inferring each reasoning step. Therefore, ideally, by generating well-formed chains, LLMs will achieve high-quality reasoning trajectories from query entities to answer candidates. However, under a training-free setting, it is almost impossible to ensure the generation of such chains simply by some conventional techniques like in-context learning or prompting \cite{Wei2021FinetunedLM}, due to the obscure structural information after linearization of graphs in input, as well as the difficulties in the injection of this concept into LLMs. To achieve this purpose, we further propose \textbf{\textit{graph-aware constrained decoding}}, to impose a constraint upon LLMs' decoding process. The constraint is induced from a local and query-centric subgraph, and as the reasoning process proceeds, the subgraph progressively expands in accordance with both the principle of well-formed chains and the topology of the source KG. By restricting the scope of valid tokens as output, this hard constraint is able to strictly regularize the LLM's generation to be a well-formed chain. Combining well-formed chains and graph-aware constrained decoding together, our approach, Decoding on Graphs (\textsc{DoG}), enables LLMs to produce reasoning trajectories that are sound and faithful to the given KGs. 

We evaluate our approach on three KGQA datasets with various open-source LLMs. The experimental results demonstrate that \textsc{DoG} effectively guarantees the generation of well-formed chains, thereby leading to higher accuracy. The key contributions of this work are: (1) We define well-formed chains as a principle for faithful and sound reasoning on KGs; (2) To achieve reasoning with well-form chains, we propose graph-aware constrained decoding for LLMs; (3) The experiments show that our approach exhibits the best performances on three KGQA benchmarks under a training-free setting.


\begin{figure*}[ht]
\centering
\includegraphics[width=1\textwidth]{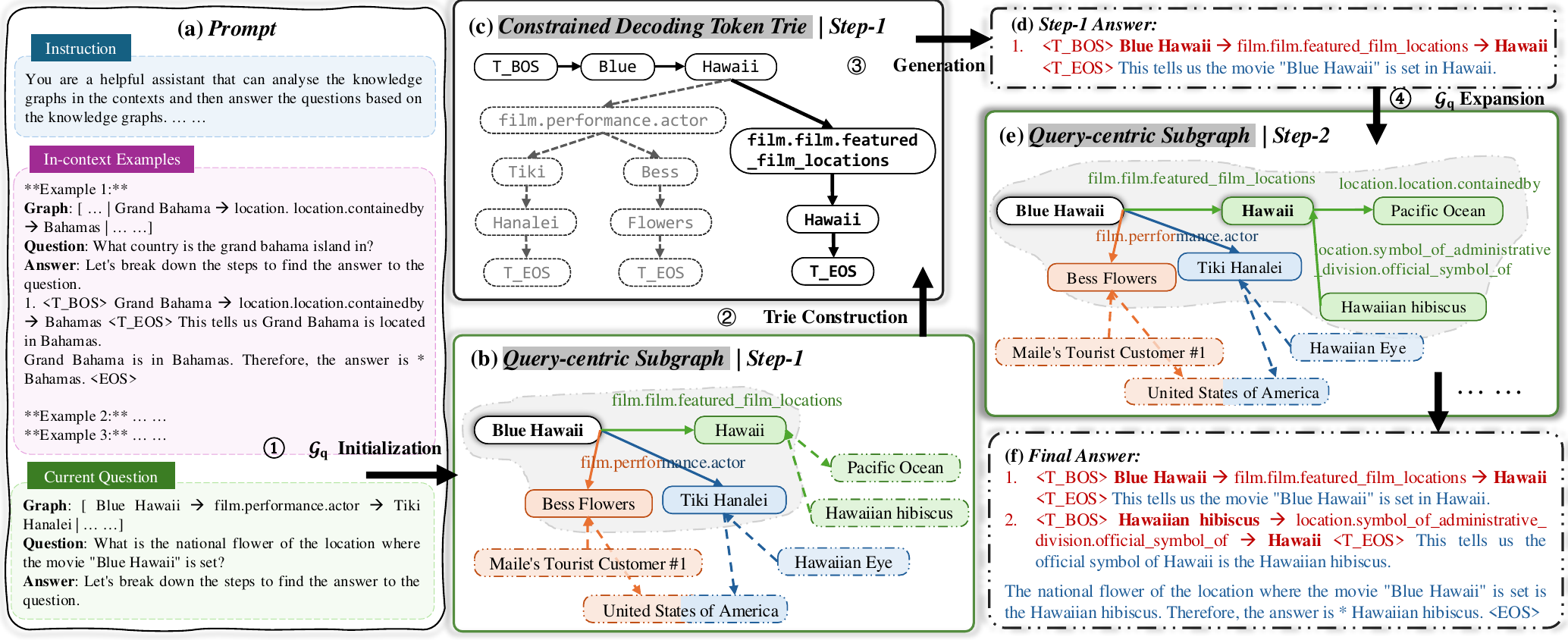}
\caption{An example workflow of \textsc{DoG} with beam size of 1. The input consists of the instruction, three in-context learning examples and the current question, with the full prompt detailed in Tab. \ref{tab:appendix-prompt}. The 2-hop input graph is illustrated in (b) and (e), including entities and relations in both solid and dotted lines. Starting from the query entity (white node), the query-centric subgraph $\mathcal{G}_q$ (grey area) is initialized by adding all triplets associated with the query entity, represented with solid lines in (b). The corresponding trie for constrained decoding is shown in (c), maintaining a set of valid tokens $w\in W_{val}$ for each position within Step-1. \textsc{DoG} chooses \texttt{(Blue Hawaii -> film.film.featured\_film\_locations -> Hawaii)} as Step-1 triplet, branch highlighted in \textbf{bold} within (c). Followed by unconstrained generation, Step-1 result is in (d) with well-formed chain in \textcolor{red}{red} and standard decoding in \textcolor{blue}{blue}. The process advances to Step-2 in (e), where all triplets outside $\mathcal{G}_q$ that involve the two visited entities with \textbf{boldface} are added.
The final answer is provided in (f).} 
\label{tab:dog-overview}
\vspace{-0.6\baselineskip}
\end{figure*}

\section{Decoding on Graphs}
\subsection{Task Formulation}

Given a KG $\mathcal{G}$ which can be represented by a set of relation triplets $t=(e, r, e^{'})$, where $e, e^{'}, r$ denote the head entity, tail entity, and corresponding relation respectively, the task of KGQA is to answer a natural language question $q$ through finding the supporting information embedded in $\mathcal{G}$. 

While prior works predominantly rely on specialized subgraph retrievers or iterative LLM-prompting to discover helpful information on KGs, we propose \textsc{DoG} to make LLMs to generate well-formed chains on KGs (§\ref{sec_chain}) by enforcing graph-aware constrained decoding (§\ref{sec_decoding}) and beam search execution (§\ref{sec_beam_search}), finally leading to better reasoning trajectories and answer candidates.

\subsection{Well-formed Chain}
\label{sec_chain}
Based on a KG $\mathcal{G}$ and a question $q$, we aim to generate a well-formed chain, which is a sequence of fact triplets $T=\{t_1, t_2, ..., t_{|T|}\}$ and would lead to an answer.\footnote{A useful and well-formed chain does not necessarily include the answer as an entity, but contains all the information necessary for deriving the answer.} Here we give the definition of well-formed chain. 
A chain of triplets is well-formed chain if
\begin{enumerate}
    \item all the triplets along the chain exist on the KG, formally, $t_i \in \mathcal{G}$ for $1 \leq i \leq |T|$; 
    \item either head or tail entity of each triplet has been visited in the previous triplets or referred to as a query entity in the question $q$. 
\end{enumerate}
A well-formed chain is desirable for reasoning of KGQA in that (1) hallucination can be avoided during the reasoning process due to Property 1; (2) the search scopes for planning each reasoning step can be narrowed down, as only the neighbors of the already visited entities instead of the whole KG should be explored, according to Property 2.

We use in-context learning \cite{Wei2021FinetunedLM} for guiding the LLM to output a chain of triplets (we do not say well-formed chain here, as it is difficult to achieve this solely using in-context learning). The prompt that contains three in-context examples with the desired output format is demonstrated in Tab. \ref{tab:appendix-prompt}. 
We transform the whole graph into a linearized form to accommodate text-form input.

Nevertheless, it is difficult to ensure the generation of well-formed chains by solely using in-context learning (§\ref{sec_path_acc_analysis}). This is because, given the linearized form of a large-sized KG, LLMs may struggle to comprehend the structure of the graph. The challenge will be even more pronounced when a chain of multiple triplets is required to solve multi-hop questions. Next, we apply graph-aware constrained decoding to open-source LLMs to guarantee the generation of well-formed chains.

\subsection{Graph-Aware Constrained Decoding}
\label{sec_decoding}
As revealed by \citet{Wang2024ChainofThoughtRW}, chain of thoughts (CoT) reasoning paths can be effectively elicited from LLMs by simply altering the decoding process. Inspired by this, \textsc{DoG} regulates the decoding process of LLMs with KG topology as a constraint for generating well-formed chains.

\noindent{\textbf{Query-centric Subgraph}} To effectively model the constraint, for question $q$, we maintain a local and query-centric subgraph $\mathcal{G}_q$ throughout the reasoning process. This subgraph is initialized as the set of triplets that contain the query entity $e_q$, formally, $\mathcal{G}_q=\{(e, r, e^{'})\mid e_q\in\{e, e^{'}\}\}$. At $i$-th step, only those triplets on the query-centric subgraph, $t \in \mathcal{G}_q$, are allowed to be generated.
Moreover, as the reasoning process proceeds, the subgraphs gradually expands in alignment with both the principle of well-formed chains and the topology of the source KG. Specifically, once $t_i=(e_i, r_i, e_i^{'})$ is generated as the $i$-th step result (as the LLM considers $t_i$ is more possible than any other triplets on $\mathcal{G}_q$), all the triplets outside $\mathcal{G}_q$ that have $e_i$/$e^{'}_{i}$ as head/tail entity, will be incorporated into $\mathcal{G}_q$, as.
\begin{equation}
\mathcal{G}_q \leftarrow \mathcal{G}_q \cup \{(e, r, e^{'})\in \mathcal{G}\mid \{e, e^{'}\} \cap \{e_i, e_i^{'}\} \neq \emptyset\}.
\label{eq_graph_update}
\end{equation}
Such a progressive expansion is essential, because it enables $\mathcal{G}_q$ to precisely encompass all the possibilities of the next triplet as the chain continues. An example is shown in Fig. \ref{tab:dog-overview} (b) to (c).

It is easy to infer that constraining the generation of each reasoning step to the triplets on $\mathcal{G}_q$ can ensure the well-formedness of generated triplet chains. Please note that query-centric subgraph described here are not the graphs used in input. The input incorporates a complete KG, of which the query-centric subgraph is a subset.

\noindent{\textbf{Constrained Decoding}}
To force the output of each step to be a triplet on $\mathcal{G}_q$, we impose a dynamic constraint on the output vocabulary during LLMs' decoding, and this constraint only allows the output of those valid tokens $w\in W_{val}$ which can constitute a triplet on $\mathcal{G}_q$. However, the set of valid tokens $W_{val}$ would change as the generation continues. To update this constraint on-the-fly, we use a trie to keep track of $W_{val}$. For example, at the first step, the trie is constructed as in Fig.\ref{tab:dog-overview}(c) based on $\mathcal{G}_q$ in Fig.\ref{tab:dog-overview}(b). When the text generated so far within the first step is ``\textit{Blue Hawaii}'', we have $W_{val}$ = \{``\textit{film.performance.actor}'', ``\textit{film.film.featured\_film\_location}''\}, indicated by the child nodes of ``\textit{Hawaii}''\footnote{For the sake of simplicity in the demonstration, sub-tokens is not considered here.}. $W_{val}$ is updated continuously according to the path from the root to the leaves.
See App.\ref{sec:appendix_constrained_decoding} for more details on our implementation of the trie.

When generating the $i$-th token at each reasoning step, given $W_{val}$, the constraint is executed with modifying the logits of the output vocabulary as
\begin{equation}
logit_{w}=\left\{\begin{array}{ll}
logit_{w},  &w \in W_{val},\\
-\infty \quad, & \text{otherwise},
\end{array}\right.
\label{eq_logit_modification}
\end{equation}
where $logit_{w}$ represents token $w$'s logit score predicted by the LLM, which is then fed into a softmax function to obtain the final output distribution. Noteworthily, we do not modify the parameters or operating mechanism of LLMs, instead, we merely exclude those invalid generation. Therefore, the pre-trained LLMs' reasoning ability can be preserved, and is then regularized by the topology of KGs to actively uncover the possible well-formed chains. For instance, in the case above, between the two available tokens in $W_{val}$, the LLM is more likely to generate ``\textit{film.film.featured\_film\_location}'' which is more relevant to the question.

On the other hand, unconstrained generation for intermediate analyses and final conclusion, \textit{e.g.}, the text in blue in Fig. \ref{tab:dog-overview} (f), is also necessary. Therefore, the decoding constraint is lifted once a complete triplet is generated, allowing unconstrained generation of these content; When $<T\_BOS>$ (placeholder as the beginning of a triplet) is generated during the unconstrained generation, suggesting the LLM considers exploring the next triplet, the decoding constraint will take effect again; The generation of $<EOS>$ during the unconstrained generation will terminate the whole reasoning.

\subsection{Beam Search for \textsc{DoG}}
\label{sec_beam_search}

When generating a triplet chain, a vanilla practice is to produce only one triplet for each step\footnote{After the logit manipulation in Eq.\ref{eq_logit_modification}, greedy search, beam search or other sampling methods can be used to sample tokens}. This however may cause error propagation --- inferring an irrelevant (though valid) triplet will negatively affect subsequent reasoning. To alleviate this problem, we further integrate \textsc{DoG} with triplet-level beam search execution, to make it possible to consider multiple valid triplets at each reasoning step. 

The complete approach is formalized in Algo. \ref{algo}, involving a token-level beam search and a triplet-level beam search. We use beam search as the token-level sampling method as it can return multiple sequences (triplets here)\footnote{One can also use sampling method like top-k/p \cite{Holtzman2020The} for the same purpose, but greedy search is not allowed here as it only returns at most one sequence.}. The triplet-level beam search corresponds to the selection of triplet chains. Specifically, for each reasoning step, it retains $bs$ (the hyperparameter for beam search) triplet chains with the highest chain scores. The chain score $S$, which measures the confidence of a chain, is the sum of the triplet scores of all the triplets along the chain. The triplet score $s_t$ can be derived as
\begin{equation}
s_t=\log P_{\text{LLM}}(t|Q), 
\end{equation}
where $Q$ denotes the so-far generation, including the input prompt and previous reasoning steps.

With beam search execution, for each reasoning step, \textsc{DoG} is able to explore multiple triplets in parallel and then retain the most plausible chains.
\section{Experiment}
\subsection{Datasets and Knowledge Graphs} We evaluate \textsc{DoG} for KGQA on three benchmarks: WebQuestionSP (\textbf{WebQSP}, \citeauthor{yih-etal-2016-value}, \citeyear{yih-etal-2016-value}), Complex WebQuestion (\textbf{CWQ}, \citeauthor{talmor-berant-2018-web}, \citeyear{talmor-berant-2018-web}) and \textbf{2Wikimultihop} \cite{ho-etal-2020-constructing}. WebQSP and CWQ are two widely-used KGQA benchmarks with Freebase \cite{10.1145/1376616.1376746} as the background knowledge graph, containing up to 2-hop and 4-hop questions respectively. For each question in WebQSP, we extract all the entities within a 2-hop distance from the query entities on Freebase, and these entities along with the corresponding relations will constitute a KG. For CWQ, we do this with 4-hop distance. However, the resulting KGs are prohibitively large. We further utilize a lightweight text embedding model,\texttt{stella\_en\_400M\_v5} \footnote{\url{https://huggingface.co/dunzhang/stella_en_400M_v5}}, to rank the triplets based on their semantical similarities to the questions, and finally retain the top-120 triplets, leading to the final KGs used for evaluation. 

2Wikimultihop is a more challenging benchmark with a large proportion of multi-hop questions that require different types of reasoning, like comparison and composition. Besides, some instances on the dataset offer the annotation of the ground truth reasoning path, allowing the assessment on the generated triplet chains. For each instance, 10 relevant passages are provided as context information. Following \citet{li-du-2023-leveraging} and \citet{Fang2024TRACETE}, we instruct an LLM, \texttt{Gemma-2-9b-it} \cite{Riviere2024Gemma2I}, to extract the underlying knowledge graphs from the passages with in-context learning. We do not filter the resulting graphs as they are often moderate-sized. Note that the original passages are no longer used in input when taking evaluation. See App. \ref{sec:appendix_graph_construction} for more details on the graph construction. 


\begin{table}[]
\centering
\scalebox{0.66}{
\begin{tabular}{l|ccc}
\toprule
Dataset & \# instances & \begin{tabular}[c]{@{}c@{}}average graph size \\ ( \# triplets)\end{tabular} & \begin{tabular}[c]{@{}c@{}}\% multi-hop \\ question (\textgreater{}1 hop)\end{tabular} \\ \midrule
WebQSP & 1542 & 119.71 & 2.01 \\
CWQ & 2617 & 119.64 & 28.58 \\
2Wikimultihop & 6964 & 64.58 & 100.00 \\ \bottomrule
\end{tabular}}
\caption{Data statistics of three processed datasets. The hop number of a question is measured by the length of the shortest path from the query entity to the answer question, which may be shorter than the ground truth path (which is unavailable for WebQSP, CWQ and part of 2Wikimultihop). }
\label{tab:data-statistics}
\vspace{-1\baselineskip}
\end{table}
Due to the incompleteness of the source KG and the imperfection of the graph construction process, some instances may have a graph that excludes the gold answer. Finally, we filter out these instances\footnote{For excluding these instances, we compare the names of all the entities name on the graph and the gold answer simply using text matching. However, this method can not exclude those false positive instances in which the answer entity appears on the KG but the relation paths from query entity to answer entity are not relevant to the corresponding questions.}, as we want to focus on LLMs' ability of reasoning on graphs. We found that the LLMs would rely on their internal knowledge to make a prediction if no plausible answer on the given KGs. The statistics of the processed datasets are shown in Tab. \ref{tab:data-statistics}. 

\subsection{Implementation and Evaluation Metrics}
\label{sec_implementation}
To ascertain the general applicability of our approach, we apply \textsc{DoG} to three open-source LLMs, \texttt{Llama-3.1-8B-it}\cite{dubey2024llama3herdmodels}, \texttt{Gemma-2-9b-it}\cite{Riviere2024Gemma2I} and \texttt{Qwen-2.5-7b-it}\cite{qwen2.5}, where \texttt{-it} means instruction-tuned versions. We implement the decoding method using Huggingface \cite{wolf2020huggingfacestransformersstateoftheartnatural} framework.
Notably, we refrain from fine-tuning of these models.
Following previous works \cite{sun2024thinkongraph, baek-etal-2023-knowledge, jiang-etal-2023-structgpt}, we report the performances for all benchmarks using $Hits\text{@}1$, which measures the proportion of instances whose top-1 prediction is the ground truth answer.

\begin{table*}[t]
\centering
\scalebox{0.64}{
\begin{tabular}{ll|ccc ccc ccc}
\toprule
 &  & \multicolumn{3}{c}{\textbf{Llama 3.1-8B}} & \multicolumn{3}{c}{\textbf{Gemma 2-9B}} & \multicolumn{3}{c}{\textbf{Qwen 2.5 -7B}} \\
 & Approach & WebQSP & CWQ & 2Wikimultihop & WebQSP & CWQ & 2Wikimultihop & WebQSP & CWQ & 2Wikimultihop \\ \midrule  \midrule
\multirow{2}{*}{\begin{tabular}[c]{@{}l@{}}\textit{Vanilla} \\ \textit{Baselines}\end{tabular}} & Direct Answering & 87.55 & 67.10 & 54.58 & 84.95 & 60.68 & 58.85 & 88.72 & 67.41 & 50.92 \\
 & CoT & 89.88 & 72.14 & 80.92 & 88.07 & 67.48 & 80.00 & 92.22 & 71.42 & 79.97 \\ \midrule  \midrule
\multirow{2}{*}{\begin{tabular}[c]{@{}l@{}}\textit{Iterative LLM} \\ \textit{Prompting}\end{tabular}} & ToG & 83.59 & 52.66 & 64.03 & 78.02 & 44.33 & 64.27 & 81.52 & 51.39 & 60.21 \\
 & Tree-of-Traversals & 79.18 & 53.92 & 68.42 & 84.82 & 66.54 & 75.00 & 85.78 & 57.70 & 73.41 \\ \midrule  \midrule
\multirow{2}{*}{\begin{tabular}[c]{@{}l@{}}\textit{Specialized} \\ \textit{Retrievers}\end{tabular}} & RoG & 83.98 & 64.12 & - & 81.00 & 58.20 & - & 84.50 & 65.11 & - \\
 & GNN-RAG & 87.42 & 74.89 & 42.00 & 83.53 & 68.51 & 26.78 & 86.96 & 71.38 & 37.54 \\ \midrule  \midrule
\begin{tabular}[c]{@{}l@{}}\textit{Structure}\\ \textit{Training}\end{tabular} & StructLM-Mistral$^*$ & 91.31 & 72.98 & 58.20 & - & - & - & - & - & - \\ \midrule  \midrule
\multirow{3}{*}{\textit{Ours}} & DoG(bs=1) & 91.05 & 75.55 & 83.99 & 90.27 & 70.77 & 82.30 & \textbf{92.67} & 73.75 & 82.98 \\
 & DoG(bs=2) & 90.99 & 76.08 & 83.54 & \textbf{91.57} & 72.56 & 83.49 & 92.60 & 73.71 & 83.31 \\
 & DoG(bs=3) & \textbf{91.38} & \textbf{76.16} & \textbf{84.06} & 91.37 & \textbf{74.10} & \textbf{84.06} & \textbf{92.67} & \textbf{74.17} & \textbf{84.16} \\ \bottomrule
\end{tabular}}
\caption{Performance comparison of different methods on the three KGQA benchmarks. The best performance is highlighted in \textbf{bold}. All results are obtained using the same input graphs. Scores for \texttt{StructLM-Mistral$^*$} are reproduced using the open-sourced model \cite{zhuang2024structlm} trained based on \texttt{Mistral-7B-Instruct-v0.2}. \texttt{RoG} cannot be applied on 2Wikimultihop, as its checkpoint trained on Freebase fails to generate relation labels of Wikidata, the background KG of 2Wikimultihop.
}
\label{tab:main-table}
\vspace{-1\baselineskip}
\end{table*}

\subsection{Baselines}
We compare \textsc{DoG} with four types of baselines. 
First, to validate the effectiveness of each design in \textsc{DoG}, we involve \textit{Vanilla Baselines}: (1) \textbf{Direct Answering} is achieved with in-context learning, and the prompt is similar to the one used in \textsc{DoG} except that the triplet chains are omitted to encourage LLMs to give predictions directly. Further, we ablate the graph-aware constrained decoding component in \textsc{DoG} and obtain (2) \textbf{CoT}, which can be considered as a vanilla chain-of-thought \cite{wei2022chain} method. 
Secondly, to highlight the benefits of training-free DoG, we consider two baselines that train in-domain retrievers over KGs (\textit{Specialized Retrievers}). (3) \textbf{RoG} \cite{luo2024reasoning} jointly trains LLMs for the generation of relation paths and the reasoning over valid retrieved KG paths. (4)  \textbf{GNN-RAG} \cite{mavromatis2024gnnraggraphneuralretrieval} trains a GNN retriever to select relevant KG reasoning paths. We reproduce the performance on our processed datasets using the two retrievers trained by the authors on CWQ and WebQSP. The retrieval results are then verbalized into the same LLM reasoners listed in §\ref{sec_implementation} with Direct Answering. 
Thirdly, we show the effectiveness of decoding-based \textsc{DoG} over two \textit{Iterative LLM-Prompting} approaches which heavily rely on the instruction-following abilities of LLMs and typically base their decisions on a single hop at a time. (5) \textbf{ToG} \cite{sun2024thinkongraph} treats LLMs as an agent to iteratively execute beam search on KG by exploring relevant facts hop-by-hop. It first selects significant relations and then uses selected relations to guide entity exploration. (6) \textbf{Tree-of-Traversals} \cite{markowitz-etal-2024-tree} is a zero-shot approach that augments LLMs with the interface of KGs. It repeatedly expands the local subgraph with an action state machine and tree search algorithm. We use the authors' open-source codes and their default settings for reproduction.
Lastly, we compare the structure-decoding \textsc{DoG} over the \textit{structure-training} method \textbf{StructLM} \cite{zhuang2024structlm}, which is designed to train generalist models with a large amount of structured data. The results are reproduced using the model checkpoint and prompt provided by the authors.

\section{Main Results}
\label{section_main_results}

Tab. \ref{tab:main-table} shows the results of \texttt{DoG} across three benchmarks. Some general patterns are manifest across different foundation LLMs:

\noindent{\textbf{Effectiveness of reasoning with triplet chains}} \texttt{Direct Answering} shows considerable performance on WebQSP, corroborating the findings by \citet{Dai2024LargeLM} that pre-trained LLMs possess the ability of comprehending graph-structured data to some extent. However, the performance drops dramatically for CWQ and 2Wikimultihop which have a larger proportion of multi-hop questions (see Tab. \ref{tab:data-statistics}). In comparison, \texttt{CoT} exhibits notably better performance on the two datasets, indicating the benefits brought by the reasoning processes in the form of triplet chains. 

\noindent{\textbf{Effectiveness of graph-aware constrained decoding} Even with identical input prompts, \texttt{DoG} (beam size = 1) outperforms \texttt{CoT} with notable gaps, verifying the effectiveness of graph-aware constrained decoding. The detailed analysis for the improvement is provided in §\ref{sec_path_acc_analysis}. It is also noteworthy that \texttt{DoG} surpasses \texttt{StructLM} which underwent extensive pre-training on structured data, making graph-aware constrained decoding a more affordable yet better-performing alternative to the pre-training. 

\noindent{\textbf{Effectiveness of beam search execution} \texttt{DoG} shows improving performance with larger beam sizes in most cases. The improvements are prominent on CWQ and 2Wikimultihop while WebQSP sees limited gains, implying that the beam search execution could benefit multi-hop questions more.

\noindent{\textbf{Superiority of explicit reasoning over specialized retrievers} The graph retrievers of \texttt{GNN-RAG} considerably boosts \texttt{Direct Answering}'s performances on CWQ, and even outperforms \texttt{CoT}. We found that the retriever always returns a highly concentrated subgraph which only contains a few relation paths towards the plausible answers, greatly reducing the difficulty of finding answers for LLMs. However, there are obvious degradations in the performance on 2Wikimultihop, highlighting the specially trained retrievers' inability to adapt to out-of-domain scenarios. Because the relation labels on 2Wikimultihop mainly follow the taxonomy of Wikidata, quite divergent from that of Freebase on which \texttt{GNN-RAG} and \texttt{RoG} are based. In contrast, \texttt{DoG} shows consistently enhanced performance across three benchmarks. \texttt{DoG} utilizes the pre-trained abilities of LLMs and textual information from verbalized KGs to make explicit reasoning, rendering it considerably robust to different KGQA tasks with diverse background KGs.

\noindent{\textbf{Superiority of direct generation over iterative prompting} Although both \texttt{DoG}  and iterative LLM-prompting are training-free methods, \texttt{DoG} shows superior performance, especially on CWQ and 2Wikimultihop. Iterative prompting places high demands on the LLMs' abilities, as our manual check found that the LLMs struggle on handling the instructions for some complex operations over KGs. Contrarily, \texttt{DoG} finds answers through the straightforward generation of a well-defined reasoning trajectory and thereby works well with small-scaled LLMs.
Moreover, for both \texttt{ToG} and \texttt{Tree-of-Traversals}, only a very small portion of the grounded KG is visible to the LLMs when deciding operations over the KG, which may bring adverse effects to the path-planning process for finding answers. More discussion on this is in §\ref{sec_visibility_analysis}.


\section{Analysis}
\subsection{Effect of Graph-Aware Constrained Decoding}
\label{sec_path_acc_analysis}
\begin{figure}[ht]
\centering
\includegraphics[width=0.35\textwidth]{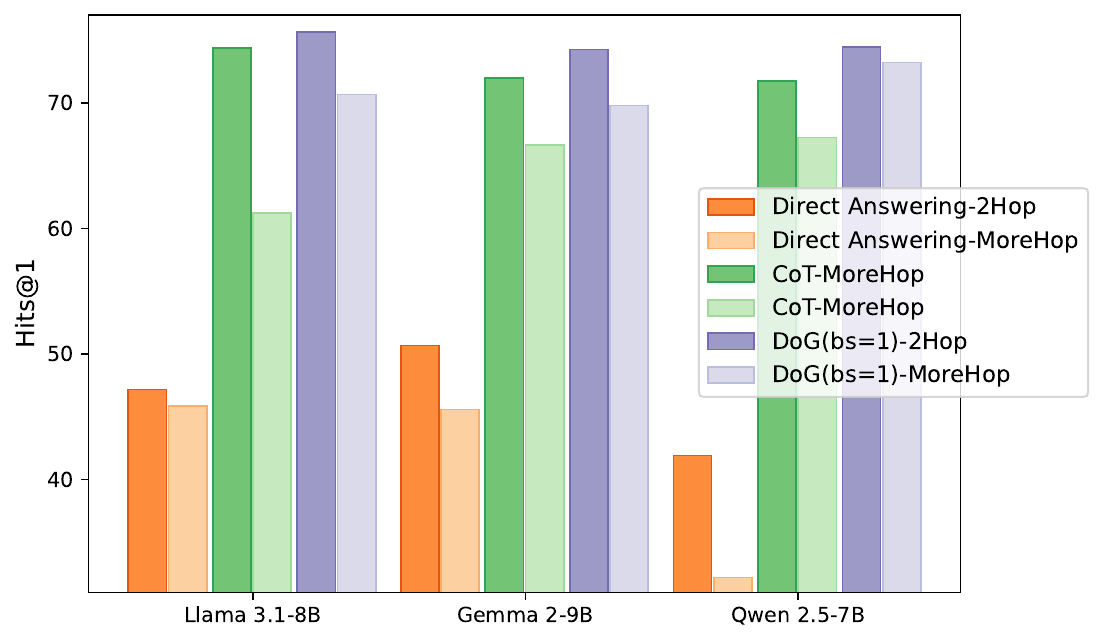}
\caption{Performance of \texttt{Direct Answering}, \texttt{CoT} and \texttt{DoG} (beam size = 1) on 2Wikimultihop with 2-hop and >2 hop instances.}
\label{fig:analysis-5.1}
\vspace{-0.5\baselineskip}
\end{figure}


\begin{table}[]
\centering
\scalebox{0.68}{
\begin{tabular}{lccccc}
\toprule
\multicolumn{6}{c}{\textit{(a) Triplet-F1 $\uparrow$}} \\ \midrule
 & \multicolumn{2}{c}{\textbf{CoT}} & \multicolumn{2}{c}{\textbf{DoG(bs=1)}} & \textbf{Ground-Truth} \\
Model & 2-hop & $>$2-hop & 2-hop & $>$2-hop & $\ge$2-hop \\ \midrule
\multicolumn{1}{l|}{Llama} & 75.77 & 66.80 & 79.85 & \multicolumn{1}{c|}{79.93} & 100.00 \\
\multicolumn{1}{l|}{Gemma} & 67.39 & 68.07 & 70.73 & \multicolumn{1}{c|}{68.79} & 100.00 \\
\multicolumn{1}{l|}{Qwen} & 75.35 & 66.19 & 80.41 & \multicolumn{1}{c|}{74.61} & 100.00 \\ \midrule \midrule
\multicolumn{6}{c}{\textit{(b) \% ill triplet $\downarrow$}} \\ \midrule
 & \multicolumn{2}{c}{\textbf{CoT}} & \multicolumn{2}{c}{\textbf{DoG(bs=1)}} & \textbf{Ground-Truth} \\
Model & 2-hop & $>$2-hop & 2-hop & $>$2-hop & $\ge$2-hop \\ \midrule
\multicolumn{1}{l|}{Llama} & 13.44 & 16.67 & 0.00 & \multicolumn{1}{c|}{0.00} & 0.00 \\
\multicolumn{1}{l|}{Gemma} & 13.41 & 11.26 & 0.00 & \multicolumn{1}{c|}{0.00} & 0.00 \\
\multicolumn{1}{l|}{Qwen} & 12.64 & 19.81 & 0.00 & \multicolumn{1}{c|}{0.00} & 0.00 \\ \bottomrule
\end{tabular}
}
\caption{Analysis of \texttt{CoT} and \texttt{DoG} (beam size = 1) on 2Wikimultihop for (a) Triplet-F1 and (b) \% ill triplet.}
\label{tab:analysis-5.1-triplet-details}
\vspace{-1\baselineskip}
\end{table}
\begin{table*}[]
\centering
\scalebox{0.71}{
\begin{tabular}{l|ccc|ccc|ccc}
\toprule
 & \multicolumn{3}{c|}{\textbf{Llama 3.1-8B}} & \multicolumn{3}{c|}{\textbf{Gemma 2-9B}} & \multicolumn{3}{c}{\textbf{Qwen 2.5 -7B}} \\
 & WebQSP & CWQ & 2Wikimultihop & WebQSP & CWQ & 2Wikimultihop & WebQSP & CWQ & 2Wikimultihop \\ \midrule
\multicolumn{1}{l|}{DoG w/ local visibility} & 91.05 & 74.70 & 80.00 & 90.40 & 72.30 & 82.06 & 86.32 & 68.48 & 76.22  \\
\multicolumn{1}{l|}{DoG w/ full visibility} & 90.99 & 76.08 & 83.54 & 91.57 & 72.56 & 83.49 & 92.60 & 73.71 & 83.31 \\ \bottomrule
\end{tabular}}
\caption{Performance of \texttt{DoG} (beam size = 2) with different levels of graph visibility.}
\label{tab:analysis-5.2-visibility}
\vspace{-0.7\baselineskip}
\end{table*}
\begin{table}[]
\centering
\scalebox{0.71}{
\begin{tabular}{l|cc|cc}
\toprule
 & \multicolumn{2}{c|}{\textbf{WebQSP}} & \multicolumn{2}{c}{\textbf{CWQ}} \\
Model & DoG & DoG + GNN & DoG & DoG + GNN \\ \midrule
Llama & 90.99 & 90.66 & 76.08 & 84.72 \\
Gemma & 91.57 & 89.56 & 72.56 & 80.09 \\
Qwen & 92.60 & 90.14 & 73.71 & 84.37 \\ \bottomrule
\end{tabular}}
\caption{Performance of \texttt{DoG} (beam size = 2) with the integration of specialized retriever released by \texttt{GNN-RAG} \cite{mavromatis2024gnnraggraphneuralretrieval}.}
\label{tab:analysis-5.3-retriever}
\vspace{-1\baselineskip}
\end{table}


2Wikimultihop dataset provides the annotation of ground truth reasoning paths for some of the instances, with each step represented as a relation triplet. There are 4,843 such instances on our processed dataset. We split those instances into two parts based on the number of reasoning steps\footnote{The first part contains the questions with 2 hops, the second one is with >2 hops. We do not further split the second path in a finer-grained scheme as there is only a small number of >3-hop questions.}. The evaluation on both parts for \texttt{Direct Answering}, \texttt{CoT} and \texttt{DoG} (The beam size is 1 for a fair comparison) is shown in Fig. \ref{fig:analysis-5.1}. 


As shown in Fig. \ref{fig:analysis-5.1}, \texttt{DoG} demonstrates more pronounced improvements over \texttt{CoT} for >2-hop questions, compared to 2-hop questions. For an in-depth investigation on the gaps, we examine the accuracy of the triplet chains predicted by both methods. Here we define two metrics: (1) $Triplet\!-\!F1$ which represents the F1-score between predicted triplets and ground-truth triplets, (2) $\% \ ill \ triplet$ which measures the percentage of predicted triplets that break the well-formedness of a chain (§\ref{sec_chain}). As shown in Tab.
\ref{tab:analysis-5.1-triplet-details}, all the ground-truth reasoning paths on 2Wikimultihop are well-formed chains ($\%ill\ triplet\!=\!0$), implying that well-formed chains can aid in making accurate sequential reasoning on KGs. Accordingly, it can be observed that \texttt{DoG} consistently exhibits higher $Triplet\!-\!F1$ than \texttt{CoT} due to \texttt{DoG}'s zero values of $\%\ ill\ triplet$.  Moreover, for >2-hop questions, the differences in both metrics between the two methods are more significant (except for the cases of Gemma).
\texttt{DoG}'s graph-aware constrained decoding manner effectively roots out the occurrence of ill triplets for both splits of questions, and, in turn, benefits the generation of high-quality triplet chains for multi-hop questions. 

\subsection{Effect of Global Visibility of Graphs}
\label{sec_visibility_analysis}
\texttt{DoG} incorporates a complete KG as input for global visibility of the graph. In this section, we attempt to alter this setting such that for each step, only the local query-centric graph (§\ref{sec_decoding}) instead of the complete KG is fed to the LLMs. The comparisons are illustrated in Tab. \ref{tab:analysis-5.2-visibility}. With local visibility of graphs, \texttt{DoG}'s performance incurs modest degradation. The visibility of global KG structure allows the LLMs to access information long distance from query entities, which perhaps helps plan better reasoning chains at early steps. Meanwhile, the local query-centric graph would be updated as the reasoning chain proceeds, preventing the Key-Value cache from being reused. \texttt{DoG} with local visibility thus has a noticeably slower inference speed.

\subsection{Integration with Specialized Retriever}
\label{sec_integration_gnn_analysis}

We have seen that the subgraph retriever of \texttt{GNN-RAG} can enhance the performances of \texttt{Direct Answering} on CWQ. In this section, we integrate \texttt{DoG} with this retriever by feeding the retrieved subgraphs instead of the complete KGs into \texttt{DoG} (beam size = 2). As presented in Tab. \ref{tab:analysis-5.3-retriever}, with the concentrated subgraphs retrieved by \texttt{GNN-RAG}, \texttt{DoG} gets further improved on CWQ. This suggests that \texttt{DoG} sometimes could be adversely affected by noise from the complete KG and applying \texttt{DoG} to higher-quality KGs would lead to better performances. Nevertheless, training a specialized retriever is expensive, and the retriever may struggle with out-of-domain scenarios, returning subpar subgraphs.

\section{Related Works}

Previous attempts on solving knowledge graph question answering (KGQA) often involve semantic parsing \cite{lan2022complexknowledgebasequestion, yu2023decaf} that transforms questions into logical queries (e.g., SPARQL) to be executed over KG for answers \cite{Lan2020QueryGG, Sun2020SPARQASS, ye-etal-2022-rng}. Although these approaches enable accurate and interpretable question answering, ground truth logical queries are required for training and model-generated queries are often non-executable due to syntax or semantic errors, failing to produce valid answers \cite{yu2023decaf, luo2024reasoning}. 
Recently, researchers have been exploring the joint use of KGs and large language models (LLMs) for KGQA, leveraging the advanced capabilities of LLMs for enhanced reasoning \cite{wu2023retrieverewriteanswerkgtotextenhancedllms, 10697304, 10387715, he2024gretrieverretrievalaugmentedgenerationtextual}. 
Various approaches \cite{jin-etal-2024-graph, sun-etal-2024-oda} \textit{iteratively prompt} LLMs to synergize with KGs
to select relevant information step by step \cite{, 10.1145/3637528.3671460}. ToG \cite{sun2024thinkongraph} prompts LLMs to explore relevant facts hop-by-hop using beam search on KG. Tree-of-Traversals \cite{markowitz-etal-2024-tree} is a zero-shot reasoning algorithm motivated by Tree-of-Thoughts \cite{yao2023tree}.
Instead of focusing on one hop at a time, \textsc{DoG} uses the entire question graph as input, enabling reasoning over complete knowledge.
Additionally, these prompting approaches heavily rely on the instruction-following abilities of LLMs, 
posing challenges when applied to compact LLMs. 
Some studies \cite{he2024gretrieverretrievalaugmentedgenerationtextual} focus on training \textit{specialized retrievers} to extract key information from KG and verbalize the retrieved knowledge as input to LLMs. RoG \cite{luo2024reasoning} and GNN-RAG \cite{mavromatis2024gnnraggraphneuralretrieval} train in-domain LLM and GNN retrievers respectively. 
StructLM \cite{zhuang2024structlm}, on the other hand, trains generalist models over large amount of structured data to augment the structured knowledge grounding capabilities of LLMs. 
However, fine-tuning large scale models is computationally expensive, and specialized training suffers from limited generalization capabilities. 
Conversely, \textsc{DoG} employs a general graph-aware decoding approach, eliminating the need for external retrievers or specialized training.
\section{Conclusion}
We present \textsc{DoG}, a framework that integrates LLMs' reasoning capabilities with KGs' structural knowledge in a tightly coupled manner.
The graph-aware decoding component of \textsc{DoG} effectively regulates the decoding process of LLMs with the topology of the KG, to enable the generation of well-formed chains, thereby leading to reasoning trajectories that are both sound and faithful to the KG. Experimental results demonstrate that \textsc{DoG} outperforms existing methods based on subgraph retrieval or iterative prompting and exhibits consistently robust performances across different KGs.

\section*{Limitations}
While \textsc{DoG} demonstrates notable performance on KGQA task through training-free graph-aware constrained decoding, there are some limitations to consider. First, unlike specialized subgraph retrievers and iterative LLM-based prompting approaches, \textsc{DoG} processes the entire question graph as input to facilitate faithful and sound reasoning over knowledge graphs (KGs) LLMs directly by LLMs. This, however, requires a larger context window in LLMs to accommodate both the graph and in-context examples. Secondly, although \textsc{DoG} requests the same number of forward passes as the vanilla chain-of-thoughts (CoT) prompting due to the use of Key-Value cache, it is slower in practice. This slowdown occurs because the query-centric subgraph must to be updated after each reasoning step. Further work on engineering solutions could be implemented to enhance the efficiency of LLMs when reasoning over KGs. Lastly, our experiments and evaluations have been limited to English-based benchmarks. To ensure the robustness and reliability, it is important to extend these assessments to other languages, as this will help identify any language-specific issues or inconsistencies.

\section*{Ethics Statement}
The proposed method, DoG, is a general graph-aware decoding approach designed to enhance reasoning capabilities over public knowledge graphs (KGs), thereby reducing the reliance on the prompting engineering of LLMs. While we do not expect \textsc{DoG} iteself to introduce new areas of risk, it is essential to consider the broader ethical impacts and acknowledge the potential risks associated with existing pre-trained large language models (LLMs) and KGs. (1) Both LLMs and KGs may inherently contain social biases or factual inaccuracies due to the nature of how they are constructed from real-world data. There is a risk that \textsc{DoG} may generate biased or incorrect outputs with such backbone models for KG reasoning. (2) We have not conducted an extensive safety analysis to assess the performance of \textsc{DoG} under conditions involving misleading or deceptive knowledge graphs. 

\bibliography{custom}

\appendix

\section{Prompts}
\label{sec:appendix}
\begin{table*}
\small
    \centering
    \scalebox{0.85}{
    \colorbox{purple!8}{
    \begin{tabular}{@{}p{17.2cm}}
You are a helpful assistant that can analyse the knowledge graphs in the contexts and then answer the questions based on the knowledge graphs.
\\
The answers should give the grounded reasoning chains and think step by step, and the reasoning chains should be logically complete but have as fewer steps as possible. Do not include information irrelvant to the question.
\\ \\ \\
**Example 1:**
\\ \\
Context: [ Bahamas -> location.country.first\_level\_divisions -> Grand Cay | Grand Bahama -> location.location.containedby -> Bahamas | Bahamas -> location.location.contains -> Grand Cay | Bahamas -> location.location.contains -> Grand Bahama | Grand Cay -> location.location.containedby -> Bahamas | Bahamas -> location.country.first\_level\_divisions -> East Grand Bahama | Bahamas -> location.country.first\_level\_divisions -> West Grand Bahama | Grand Bahama -> location.location.contains -> Grand Bahama International Airport | Bahamas -> location.location.contains -> East Grand Bahama | Bahamas -> location.location.contains -> West Grand Bahama | East Grand Bahama -> location.location.containedby -> Bahamas | Bahamas -> location.location.contains -> Grand Bahama International Airport | Grand Bahama -> location.location.people\_born\_here -> Hubert Ingraham | Grand Cay -> location.administrative\_division.first\_level\_division\_of -> Bahamas | Bahamas -> location.country.administrative\_divisions -> Cat Island, Bahamas | Bahamas -> location.country.administrative\_divisions -> Long Island | West Grand Bahama -> location.location.containedby -> Bahamas | Bahamas -> location.country.capital -> Nassau | Bahamas -> location.country.administrative\_divisions -> Inagua | Bahamas -> location.country.administrative\_divisions -> Exuma | Grand Bahama International Airport -> location.location.containedby -> Bahamas | Grand Bahama -> location.location.people\_born\_here -> Juan Lewis | Grand Bahama -> location.location.contains -> West End Airport ]
\\ \\
Question: What country is the grand bahama island in?
\\ \\
Answer: Let's break down the steps to find the answer to the question.
\\ \\
1. < Grand Bahama -> location.location.containedby -> Bahamas > This tells us Grand Bahama is located in Bahamas.
\\ \\ 
Grand Bahama is in Bahamas. Therefore, the answer is * Bahamas.
\\ \\  \\
**Example 1:**
\\ \\
Context: [ William Shakespeare -> people.person.profession -> Playwright | William Shakespeare -> people.person.profession -> Poet | William Shakespeare -> base.kwebbase.kwtopic.has\_sentences -> By the time these works were published in 1609, Shakespeare was an acknowledged master of drama and an established country gentleman. | William Shakespeare -> people.person.profession -> Actor | William Shakespeare -> people.person.profession -> Author | William Shakespeare -> people.person.profession -> Lyricist | In the 21 years between 1592 and 1613, Shakespeare produced more than 30 plays. -> base.kwebbase.kwsentence.previous\_sentence -> Above all, his humanity spanned all classes and circumstances ]
\\ \\
Question: What did William Shakespeare do for a living?
\\ \\
Answer: Let's break down the steps to find the answer to the question.
\\ \\
1. < William Shakespeare -> people.person.profession -> Playwright > This tells us William Shakespeare is was playwright. 
\\
2. < William Shakespeare -> people.person.profession -> Poet > This tells us William Shakespeare was a poet.
\\ \\
William Shakespeare was a playwright, and poet. Therefore, the answer is * playwright, and * poet.
\\ \\ \\
**Example 3:**
\\ \\
Context: [ Carlton the Bear -> sports.mascot.team -> Toronto Maple Leafs | Toronto Maple Leafs -> sports.sports\_team.team\_mascot -> Carlton the Bear | Carlton the Bear -> common.topic.notable\_types -> Mascot | Mascot -> type.type.properties -> Team | Toronto Maple Leafs -> sports.sports\_team.previously\_known\_as -> Toronto St. Patricks | Team -> type.property.master\_property -> Team Mascot | Toronto Maple Leafs -> sports.sports\_team.previously\_known\_as -> Toronto Arenas | m.0crt465 -> sports.sports\_league\_participation.team -> Toronto Maple Leafs | Toronto St. Patricks -> sports.defunct\_sports\_team.later\_known\_as -> Toronto Maple Leafs | Toronto Maple Leafs -> sports.sports\_team.sport -> Ice Hockey | Toronto St. Patricks -> sports.sports\_team.sport -> Ice Hockey | Toronto Arenas -> sports.defunct\_sports\_team.later\_known\_as -> Toronto Maple Leafs | Toronto -> sports.sports\_team\_location.teams -> Toronto Maple Leafs | Toronto Maple Leafs -> sports.sports\_team.location -> Toronto ]
\\ \\
Question: What is the sport played by the team with a mascot known as Carlton the Bear?
\\ \\
Answer: Let's break down the steps to find the answer to the question.
\\ \\
1. < Carlton the Bear -> sports.mascot.team -> Toronto Maple Leafs > This tells us Carlton the Bear is the mascot of the team Toronto Maple Leafs.
\\
2. < Toronto Maple Leafs -> sports.sports\_team.sport -> Ice Hockey > This tells us Toronto Maple Leafs plays Ice Hockey.
\\ \\
Carlton the Bear is the mascot of the team Toronto Maple Leafs which plays Ice Hockey. Therefore, the answer is * Ice Hockey.
\\ \\ \\
**Example 4:**
\\ \\
Context: [ \{\texttt{graph}\} ]
\\ \\
Question: \{\texttt{question}\}
\\  \\
Answer: Let's break down the steps to find the answer to the question.
\end{tabular}
}}
    \caption{Prompt for \texttt{DoG} on KGQA task. Three in-context examples with the desired output format are listed. 
}
    \label{tab:appendix-prompt}
\end{table*}
Tab. \ref{tab:appendix-prompt} presents the complete prompt used for DoG over three KGQA benchmarks. Three in-context learning examples with the desired output format are listed.

\section{Graph Construction}
\label{sec:appendix_graph_construction}
For Complex WebQuestion (CWQ, \citeauthor{talmor-berant-2018-web}, \citeyear{talmor-berant-2018-web}) and WebQuestionSP (WebQSP, \citeauthor{yih-etal-2016-value}, \citeyear{yih-etal-2016-value}), we use the Freebase data provide by \citet{He-WSDM-2021} as the source graphs. There are CVT nodes \cite{10.1007/978-3-031-47243-5_7} in Freebase knowledge graph for modeling n-ary relationships, the CVT nodes do not have real meaning. To remove CVT nodes on our dataset, we preprocess the KG to convert n-ary relationships to binary relationships by concatenating the edge labels by "-" . 
For each example, we apply a text embedding model, \texttt{stella\_en\_400M\_v5} \footnote{\url{https://huggingface.co/dunzhang/stella_en_400M_v5}} to encode the question and all the triplet in the source graph into embeddings respectively. The input of a triplet is in the form of (\{\texttt{head\_entity}\}, \{\texttt{relation}\}, \{\texttt{tail\_entity}\}). We sort all the triplets in descending  order of cosine similarity with the question embedding. A new graph $G$ is initialized as an empty set, and we gradually add the top-ranked triplet to this new graph. To maintain the connectivity of $G$, each time a top-ranked triplet $t$ is added, all the intermediate triplets along the path from the query entity to the head/tail entity of $t$ are also added to $G$. This process repeats until the size of $G$ reaches 120. 

2Wikimultihop dataset provides 10 context passages for each question. We first utilize an entity linking tool, \texttt{relik-entity-linking-large} \cite{orlando-etal-2024-relik} \footnote{\url{https://huggingface.co/sapienzanlp/relik-entity-linking-large}}, to identify distinct entities mentioned in the passages and link them to a unique identifier in Wikidata. Given the passage and all the entities identified in the passage, we instruct an LLM, \texttt{Gemma-2-9b-it}\cite{Riviere2024Gemma2I} to extract all the relation triplets using in-context learning \cite{Wei2021FinetunedLM}. The prompt is detailed in Tab. \ref{tab:appendix-prompt2}. Finally, all the extracted triplets from different passages are combined into the final graph, with no ranking or filtering applied to the triplets.
\begin{table*}
\small
    \centering
    \scalebox{0.85}{
    \colorbox{green!6}{
    \begin{tabular}{@{}p{17.2cm}}
Given the documents and some entities within the documents, extract all the relation triplets between any pairs of the entities.
\\ \\ \\ 
**Document 1:**\\ 
Title: The Return of Dr. Fu Manchu\\ 
The Return of Dr. Fu Manchu is a 1930 American pre-Code film directed by Rowland V. Lee. It is the second of three films starring Warner Oland as the fiendish Fu Manchu, who returns from apparent death in the previous film," The Mysterious Dr. Fu Manchu"( 1929), to seek revenge on those he holds responsible for the death of his wife and child.\\ \\
Entities:The Return of Dr. Fu Manchu\texttt{\textbackslash n}1930\texttt{\textbackslash n}United States\texttt{\textbackslash n}Pre-Code Hollywood\texttt{\textbackslash n}Rowland V. Lee\texttt{\textbackslash n}Warner Oland\texttt{\textbackslash n}Fu Manchu\texttt{\textbackslash n}The Mysterious Dr. Fu Manchu\texttt{\textbackslash n}1929\\\\
Relation triplets: The Return of Dr. Fu Manchu->country->United States\texttt{\textbackslash n}The Return of Dr. Fu Manchu->director->Rowland V. Lee\texttt{\textbackslash n}The Return of Dr. Fu Manchu->movement->Pre-Code Hollywood\texttt{\textbackslash n}The Return of Dr. Fu Manchu->publication date->1930\texttt{\textbackslash n}The Return of Dr. Fu Manchu->cast member->Warner Oland\texttt{\textbackslash n}The Vengeance of Fu Manchu->cast member->Warner Oland\texttt{\textbackslash n}The Mysterious Dr. Fu Manchu->cast member->Warner Oland\texttt{\textbackslash n}The Mysterious Dr. Fu Manchu->country->United States\texttt{\textbackslash n}The Mysterious Dr. Fu Manchu->publication date->1929
\\ \\ \\ 
**Document 2:**\\ 
Title: Now, Voyager\\
Now, Voyager is a 1942 American drama film starring Bette Davis, Paul Henreid, and Claude Rains, and directed by Irving Rapper. The screenplay by Casey Robinson is based on the 1941 novel of the same name by Olive Higgins Prouty. Prouty borrowed her title from the Walt Whitman poem" The Untold Want", which reads in its entirety, In 2007," Now, Voyager" was selected for preservation in the United States National Film Registry by the Library of Congress as being" culturally, historically, or aesthetically significant." The film ranks number 23 on" AFI's 100 Years ... 100 Passions", a list of the top love stories in American cinema. Film critic Steven Jay Schneider suggests the film continues to be remembered due not only to its star power, but also the" emotional crescendos" engendered in the storyline.\\ \\
Entities: "1942\texttt{\textbackslash n}United States\texttt{\textbackslash n}Drama (film and television)\texttt{\textbackslash n}Bette Davis\texttt{\textbackslash n}Paul Henreid\texttt{\textbackslash n}Claude Rains\texttt{\textbackslash n}Irving Rapper\texttt{\textbackslash n}Casey Robinson\texttt{\textbackslash n}1941\texttt{\textbackslash n}Now, Voyager (novel)\texttt{\textbackslash n}Olive Higgins Prouty\texttt{\textbackslash n}L. Fletcher Prouty\texttt{\textbackslash n}Walt Whitman\texttt{\textbackslash n}2007\texttt{\textbackslash n}Now, Voyager\texttt{\textbackslash n}National Film Registry\texttt{\textbackslash n}Library of Congress\texttt{\textbackslash n}City Lights\texttt{\textbackslash n}American Film Institute\texttt{\textbackslash n}AFI's 100 Years ... 100 Passions\texttt{\textbackslash n}Steven Jay Schneider"\\\\
Relation triplets: Now, Voyager->country->United States\texttt{\textbackslash n}Now, Voyager->director->Irving Rapper\texttt{\textbackslash n}Now, Voyager->genre->Drama (film and television)\texttt{\textbackslash n}Now, Voyager->publication date->1942\texttt{\textbackslash n}Now, Voyager->cast member->Bette Davis\texttt{\textbackslash n}Now, Voyager->cast member->Paul Henreid\texttt{\textbackslash n}Now, Voyager->cast member->Claude Rains\texttt{\textbackslash n}Now, Voyager->screenplay writer->Casey Robinson\texttt{\textbackslash n}Now, Voyager->source material->Now, Voyager (novel)\texttt{\textbackslash n}Now, Voyager (novel)->publication date->1941\texttt{\textbackslash n}Now, Voyager (novel)->author->Olive Higgins Prouty\texttt{\textbackslash n}Now, Voyager->preserved by->National Film Registry\texttt{\textbackslash n}Now, Voyager->ranked in->AFI's 100 Years ... 100 Passions\texttt{\textbackslash n}Now, Voyager->analyzed by->Steven Jay Schneider\texttt{\textbackslash n}National Film Registry->maintained by->Library of Congress
\\ \\ \\ 
**Document 3:**\\ 
Title: \{\texttt{Title}\}\\ 
\{\texttt{Content}\}\\ \\
Entities:\{\texttt{Entity List}\}\\\\
Relation triplets:
\end{tabular}
}}
    \caption{Prompt for relation extraction on 2Wikimultihop. 
}
    \label{tab:appendix-prompt2}
\end{table*}

\section{Implementation of the Trie for Graph-Aware Constrained Decoding}
\label{sec:appendix_constrained_decoding}
We use a dictionary structure to implement a trie. Given a query-centric subgraph (§\ref{sec_decoding}), the function \texttt{build\_trie()} in the code block of Tab. \ref{appendix:code} is used to build the dictionary. Then, as demonstrated in the function \texttt{find\_valid\_tokens()}, for each iteration of token generation, with the so-far generation at the current reasoning step as the key,  we can efficiently get the set of valid tokens by performing a lookup in the dictionary.
\onecolumn
\begin{table}[]
    \begin{lstlisting}
def build_trie(sub_graph: list[list[int]]) -> dict[int, dict]:
    # sub_graph: the list that contains the token ids of all the triplet, triplets are in the form of "<T_BOS> <Triplet> <T_EOS>".
    trie = dict()
    for triplet in sub_graph:
        for suffix in [triplet[i:] for i in range(len(triplet))]:
            node = trie
            for token in suffix:
                if token not in node:
                    node[token] = dict()
                node = node[token]

    return trie

def find_valid_tokens(trie: dict[int, dict], prefix: list[int]) -> list[int]:    
    # prefix: the so-far genetation at at the current reasoning step
    v = trie
    for k in prefix:
        v = v.get(k, {})
    return list(v.keys())

\end{lstlisting}
\caption{Code for implementation of the trie for graph-aware constrained decoding.}
\label{appendix:code}
\end{table}




\begin{algorithm*}[t]
\caption{\textsc{DoG} with beam search}
    \label{algo}
    \begin{algorithmic}[1]
    \REQUIRE Pre-trained $LLM$, knowledge graph $\mathcal{G}$, input prompt $Q$ (containing $\mathcal{G}$ and question $q$), query entity $e_q$, beam size $bs$, maximum number of steps $T$
    \STATE $\mathcal{G}_q\leftarrow\{(e, r, e^{'})\in \mathcal{G}\mid e_q\in\{e, e^{'}\}\}$ $\triangleright$ initialize query-centric subgraph
    \STATE $\mathcal{P}\leftarrow\{(Q, \mathcal{G}_q, 0)\}$ $\triangleright$ initialize candidate pool with candidates comprised of input, a query-centric subgraph and a chain score
    \FOR{$t = 1,..., T$} 
        \STATE \colorbox[HTML]{ECF4FF}{// Triplet-level beam search}
        \STATE $\mathcal{P}^{'} \leftarrow \emptyset$
        \FOR{$i = 1,..., |\mathcal{P}|$}
            \STATE $Q, \mathcal{G}_q, S\leftarrow \mathcal{P}_i$ 
            \STATE $(r_1, s_1), ..., (r_{bs}, s_{bs}) \sim LLM(\cdot |Q, \mathcal{G}_q)$ $\triangleright$ sample $bs$ triplets along with the triplet scores from $LLM$ based on input $Q$ and constraint $\mathcal{G}_q$, using \colorbox[HTML]{ECF4FF}{token-level beam search}
            \FOR{$j = 1,..., bs$}
                \STATE $\mathcal{G}_q^{'}, \leftarrow Update(\mathcal{G}_q, \mathcal{G}, r_j)$ $\triangleright$ update $\mathcal{G}_q$ for $r_j$ as in Eq. \ref{eq_graph_update}
                \STATE $\mathcal{P}^{'} \leftarrow \mathcal{P}^{'} \cup \{(concate(Q, r_j), \mathcal{G}_q^{'}, S+s_j)\}$ $\triangleright$ update the input with the latest reasoning step, and add the triplet score to the chain score
            \ENDFOR
        \ENDFOR
    \STATE $\mathcal{P}\leftarrow Top(\mathcal{P}^{'} , bs)$ $\triangleright$ retain the top-$bs$ candidates in $P^{'}$ based on their chain scores
    \ENDFOR
    \RETURN $\mathcal{P}_1$
    \end{algorithmic}
\end{algorithm*}

\end{document}